\documentclass[11pt]{article}

\usepackage[]{acl}

\usepackage{times}
\usepackage{latexsym}
\usepackage{hyperref}

\usepackage[T1]{fontenc}

\usepackage[utf8]{inputenc}

\usepackage{microtype}

\usepackage{inconsolata}

\usepackage{graphicx}

\usepackage[table]{xcolor}
\usepackage{soul}

\title{Measuring Human-Like Bias in LLMs? \\A Critique of Human-Derived Bias Constructs in LLM Evaluation}

\author{
 \textbf{Antonela Tommasel\textsuperscript{1,3}},
 \textbf{Markus Schedl\textsuperscript{1,2}}
\\
 \textsuperscript{1}Johannes Kepler University Linz, Austria \\
 \textsuperscript{2}Linz Institute of Technology, Linz, Austria \\
 \textsuperscript{3}ISISTAN, CONICET-UNICEN, Argentina
 \\
\texttt{\{antonela.tommasel, markus.schedl\}@jku.at}
}

\begin{document}
\maketitle

\begin{abstract}
Researchers increasingly use human-derived bias constructs to study Large Language Models (LLMs), including social-cognitive constructs such as implicit bias and stereotype activation, and cognitive biases such as anchoring, framing effects, and confirmation bias. 
Such approaches offer alternatives to overt bias probes, particularly when direct questioning may obscure bias or when model behaviour appears normatively acceptable.  
However, adapting human bias constructs to LLMs introduces an inferential gap. Psychological instruments were developed to study human cognition and social behaviour, whereas LLM evaluations rely on probabilities, text completions, rankings, or simulated decisions. 
This paper critiques human-centered bias evaluation in LLMs. We show how this gap arises from mismatches pertaining to human-derived constructs, human-model differences, and evaluation contexts, which can blur distinct interpretations of model bias.
We then introduce a framework providing an analytical lens for relating these elements to warranted interpretations, with attention to target constructs, operationalizations, scope of inference, and limits of human analogy.
\end{abstract}

\section{Introduction}\vspace{-0.2cm}
Bias evaluation in Large Language Models (LLMs) increasingly relies on constructs originally developed to explain human cognition and social behaviour \cite{sumita2025cognitive}. Recent studies (e.g., \cite{hida-etal-2025-social}) test whether models exhibit stereotype-consistent associations, implicit-bias-like patterns, socially desirable responses, anchoring effects, framing sensitivity, confirmation bias, or demographic disparities. These approaches go beyond overt bias probes, where models are directly asked to express attitudes, judgments, or preferences, and beyond aggregate performance metrics that may obscure systematic variation across conditions. They are especially relevant when instruction tuning and safety training shape models toward normatively acceptable answers.

However, adapting human-derived bias constructs to LLMs introduces an \textit{inferential gap}\footnote{Appendix~\ref{app:existing-examples} provides examples of how this gap appears in existing LLM bias-evaluation studies.}. The gap arises from construct transfer: instruments such as implicit association tests, social desirability measures, anchoring and framing-effect tasks were developed to study human cognition, judgment, and behaviour \cite{greenwald1995implicit,greenwald1998measuring,tversky1981framing,tversky1974judgment}. 
In LLM evaluation, these constructs are operationalized through different observables, including token probabilities, generated completions, rankings, refusals, answer changes across prompts, or simulated decisions \cite{sumita2025cognitive}. 
Hence, \textit{the relationship between the psychological construct being invoked and the model behaviour being measured is not straightforward.}

\begin{figure*}[t]
    \centering
    \includegraphics[width=0.99\linewidth]{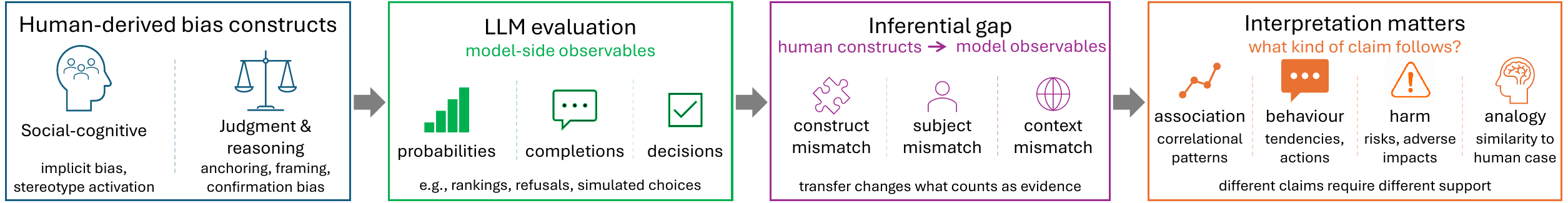}\vspace{-0.25cm}
    \caption{Conceptual overview. Human-derived bias constructs are operationalized through model-side observables in LLM evaluation, creating an inferential gap that motivates more precise interpretation of bias-related findings.}\label{fig:summary}\vspace{-0.55cm}
\end{figure*}

This paper presents a critique of human-centered bias evaluation in LLMs. We argue that the inferential gap matters because model-side evidence can support different kinds of claims.
For example, if a model assigns higher probability to \textit{man--career} than to \textit{woman--career}, this may indicate an association in the model, but it does not by itself show that the model will generate biased recommendations, cause harm in a hiring-support setting, or exhibit human-like implicit bias. 
Although our focus is on bias evaluation, similar inferential concerns may arise in other uses of psychological constructs for LLM evaluation, such as personality, emotion, moral judgment, theory of mind or pragmatic reasoning. We focus on bias because human-derived constructs are frequently used in this area and because overextended interpretations can have practical consequences for how model behaviour, harm and human analogy are communicated. 
We introduce an analytical framework that relates human-derived bias constructs, model-side observables, and warranted interpretations. \textit{The goal is not to prescribe a new evaluation protocol, but to clarify what kinds of claims different evaluation designs can support and where stronger theoretical justification is required}.  
Figure~\ref{fig:summary} provides a conceptual overview.


\vspace{-0.25cm}\section{From Evaluation of Human Bias Constructs to LLM Observables}\label{sec:evaluation-human-bias}\vspace{-0.25cm}

To make the transfer problem of human bias constructs to LLMs more concrete, we distinguish two families of LLM bias probes, 
highlighting ways in which human-derived bias constructs are translated into model-side observables\footnote{Appendix~\ref{app:bias-probe-families} provides a short example contrasting an IAT-inspired social-cognitive bias probe with an anchoring-style cognitive-bias probe.}.

\vspace{-0.25cm}\paragraph{Social-cognitive bias.} 
One line of work adapts concepts from social cognition, prejudice research, and fairness-oriented NLP to study whether LLMs reproduce socially meaningful associations or disparities. These evaluations are motivated by concerns about stereotyping, representational harms, and unequal treatment across demographic groups \cite{gallegos2024bias}. Inspired by the Implicit Association Test (IAT) \cite{greenwald1995implicit,greenwald1998measuring}, which infers relative associations from differences in response times to stimuli, recent work evaluates whether models assign higher probabilities to stereotype-consistent pairings, generate associations aligned with social stereotypes, or produce different outputs depending on demographic attributes \cite{bai2025explicitly,zhao-etal-2025-explicit}. Bias is operationalized through token likelihoods, embedding similarities, ranking preferences, controlled completions, or generation-based tasks involving demographic cues \cite{gallegos2024bias,soundararajan-delany-2024-investigating,van2025unveiling}. Related work studies socially desirable responses of LLMs, especially as instruction tuning, reinforcement learning from human feedback (RLHF), and safety alignment shape models toward normatively acceptable responses \cite{salecha2024large,ouyang2022training,bai2025explicitly}.

\vspace{-0.2cm}\paragraph{Cognitive bias in judgment and reasoning.}
A second line of work investigates whether LLMs exhibit patterns analogous to human cognitive biases in judgment and reasoning. Drawing from behavioural economics and cognitive psychology, these evaluations test whether models reproduce deviations from normative reasoning under controlled task conditions \cite{echterhoff-etal-2024-cognitive,nguyen2024human,geva2025llms}. Anchoring evaluations, for example, provide numerical anchors, prior estimates, or biased contextual information and measure whether outputs shift toward the anchor \cite{lou2026anchoring,nguyen2024human}; framing evaluations test whether logically equivalent prompts produce different answers depending on wording or presentation \cite{tversky1981framing,echterhoff-etal-2024-cognitive}. Other works study confirmation bias, overconfidence, or recency effects through reasoning tasks, decision scenarios, and prompt perturbations inspired by human experiments \cite{mina-etal-2025-cognitive,malberg2025comprehensive}.

Despite their different origins, social-cognitive and cognitive-bias evaluations share a common structure, a human-derived construct is translated into a model-side observable.
These translations make psychological paradigms useful for probing LLMs, but they also create an inferential gap. \textit{As LLMs differ from human participants in learning processes, mechanisms, and response modalities, the same construct label may refer to different kinds of evidence. }This raises questions about what kinds of interpretations model-side measurements can support \cite{cuskley2024limitations,mina-etal-2025-cognitive}.

\vspace{-0.2cm}\section{The Inferential Gap in Human-Centered Bias Evaluation}\vspace{-0.2cm}

This inferential gap is not simply a consequence of treating LLMs as another kind of human experimental subjects; this transfer changes what counts as evidence \cite{jacobs2021measurement,gawronski2022implicit}. In psychology, bias constructs are embedded in theories that connect observable responses to assumptions about cognition, behaviour, and social context \cite{greenwald1995implicit,tversky1981framing}. In LLM evaluation, construct labels are attached to outputs produced by systems with different learning processes, mechanisms, and response modalities. The relation between construct, measurement, and interpretation therefore cannot be assumed to carry over unchanged \cite{cuskley2024limitations}. 

Consider an LLM evaluation inspired by the IAT. In a human IAT \cite{greenwald1995implicit}, response-time differences are used to infer associations between groups and attributes, such as woman/man and family/career. In an LLM adaptation \cite{gawronski2022implicit}, the evaluation instead compares likelihoods of pairings such as woman-family, woman-career, man-family, and man-career, or analyses gendered continuations in occupation-related prompts. 
This preserves the broad idea of indirect association measurement, but changes the subject, the form of response being measured, and the evidential basis of the task: reaction-time differences become likelihoods, rankings or completions from a model. 
Similarly, in an anchoring probe, a shift after an arbitrary cue may indicate prompt sensitivity, task behaviour, or analogy to human anchoring, depending on how the cue, output, and construct are connected.

Drawing on validity theory in psychology and social sciences \cite{cronbach1955construct,andrade2018internal,cuskley2024limitations}, recent work framing generative AI evaluation as a social science measurement challenge \cite{wallach2025position} and validity-threat analyses in empirical software engineering \cite{wohlin2012experimentation}, we distinguish three sources of mismatch: \textit{construct}, \textit{subject}, and \textit{evaluation-context}.
These categories do not form a new validity theory; rather, they adapt familiar concerns about construct validity, external validity, and instrumentation to LLM bias evaluation.

\vspace{-0.25cm}\paragraph{Construct mismatch} concerns whether a model-side measure captures the intended theoretical construct \cite{cronbach1955construct,jacobs2021measurement}. 
Psychological bias constructs are embedded in theories connecting behaviour to assumptions about cognition, motivation, or reasoning. 
In the IAT example, the human construct is usually interpreted as an automatic association inferred indirectly from response-time differences \cite{greenwald1995implicit,gawronski2022implicit}. In an LLM adaptation, the observable may be a likelihood difference between group-attribute pairings. This may show that some associations are stronger in the model than others \cite{bai2025explicitly}. However, treating it as the same kind of implicit bias in humans requires explaining why a model probability should be interpreted as analogous to an indirect measure of automatic association.

\vspace{-0.25cm}\paragraph{Model-subject mismatch} concerns the change from human participants to computational models. 
In a human IAT, the observed response pattern is produced by participants with embodied and social experience.
LLMs are trained artifacts whose behaviour is shaped by pretraining data, fine-tuning, instruction following, RLHF, safety interventions, decoding strategies, and prompting \cite{ouyang2022training,Holtzman2020The}. 
These differences do not make comparison impossible, but limit what can be inferred from behavioural similarity alone \cite{cuskley2024limitations}. Even if humans and LLMs produce similar association patterns, they might not have the same underlying reasoning or explanation.

\vspace{-0.2cm}\paragraph{Evaluation-context mismatch} concerns how results depend on task setting, prompt design, and response modality. Human psychological tests and experiments are typically designed around specified constructs, administration conditions, response formats, scoring procedures, and validity evidence for the intended interpretations 
\cite{clark2019constructing}. 
LLM evaluations are controlled through prompts, system messages, examples, decoding, refusal policies, and conversational history, all of which can affect measured behaviour \cite{shi2024thorough,Holtzman2020The,hida-etal-2025-social}. 
In an LLM adaptation of the IAT, measured associations may vary with the prompt template, demographic labels, attribute words, examples, decoding settings, or whether the model completes text, ranks alternatives, or answers explicitly. The observed pattern therefore reflects both model behaviour and design choices in the evaluation setup, which may be hard to disentangle.

These mismatches do not imply that human-derived
constructs are inappropriate for LLM evaluation. They can provide useful structure for designing probes and interpreting model behaviour patterns. The problem arises when model-side observables are treated as if they instantiate the human construct \cite{jacobs2021measurement}. A likelihood difference, biased generation or answer shift may be meaningful evidence, but the claim it supports depends on the relation between the construct, observable behaviour, and evaluation context.

\vspace{-0.25cm}\section{An Analytical Framework for Interpreting Bias Claims}\vspace{-0.25cm}

\begin{table*}[t]
\centering
\scriptsize
\begin{tabular}{p{0.08\textwidth} p{0.22\textwidth} p{0.20\textwidth} p{0.08\textwidth} p{0.29\textwidth}}
\hline
\textbf{Interpretation} & \textbf{Claim} & \textbf{Typical model-side evidence} & \textbf{Mismatch(es)} & \textbf{Additional support needed} \\
\hline
\textbf{Distributional association} 
& The model associates some groups, attributes, cues or outcomes more strongly than others. 
& Token probabilities, likelihood contrasts, embedding similarities, association scores. 
& Construct 
& Justification that the measured association is a meaningful proxy for the intended construct. \\

\arrayrulecolor{gray}\hline

\textbf{Observable behaviour} 
& The model produces biased outputs under specific evaluation conditions. 
& Generated text, rankings, classifications, recommendations, answer shifts across prompts. 
& Subject, evaluation-context  
& Evidence that the behaviour is robust across relevant prompts, models, decoding settings or task variants. \\

\hline

\textbf{Normative harm} 
& The model behaviour may reinforce inequality, degrade decisions or cause harm in context. 
& Biased outputs connected to a use setting, affected users, groups or decisions. 
& Evaluation-context 
& An account of the deployment context, affected groups and plausible consequences. \\

\hline

\textbf{Psychological analogy} 
& The behaviour is interpreted as meaningfully similar to a human cognitive or social-cognitive bias. 
& Model behaviour plus an externally proposed mapping to a human construct. 
& Construct, subject 
& Explanation of which aspects of the human construct are preserved, absent or transformed. \\
\arrayrulecolor{black}\hline
\end{tabular}\vspace{-0.24cm}
\caption{Analytical framework for interpreting bias-related findings in LLM evaluation.}\label{tab:lens}\vspace{-0.6cm}
\end{table*}

The proposed analytical framework is inspired by validity theory in psychology and social sciences, measurement work in fairness and validity-threat analyses in empirical software engineering, which share a concern with how constructs are operationalized, how measurements support claims and how findings generalize. 

The three mismatches discussed above identify points where an evaluation result can be interpreted more strongly than the evidence warrants. 
\textit{Construct mismatch} cautions against treating model-side measures as direct instances of human constructs \cite{jacobs2021measurement}.
\textit{Subject mismatch} cautions against inferring human-like mechanisms from behavioural similarity alone \cite{cuskley2024limitations}. 
\textit{Evaluation-context mismatch} cautions against generalizing from prompt-conditioned outputs to broader claims about behaviour \cite{hida-etal-2025-social}. 
Building on these cautions, we distinguish four interpretations of bias-related findings: \textit{distributional association}, \textit{observable behaviour}, \textit{normative harm}, and \textit{psychological analogy}. 

Table~\ref{tab:lens} presents the analytical framework, linking each interpretation to typical model-side evidence, relevant mismatches, and the additional support needed to justify the claim. The framework is not intended to replace existing empirical evaluation methods with a new metric, benchmark or standardized evaluation protocol. 
Rather, it targets the inferential step connecting an empirical measurement to a bias claim. While existing practices often advise caution when interpreting LLM bias results, our framework specifies what kind of caution is needed, which type of claim is warranted by which type of evidence, and what additional support is required for stronger claims.  

\vspace{-0.25cm}\paragraph{Distributional association} refers to stronger model-side associations among groups, attributes or outcomes, measured through likelihood contrasts, embedding similarities or association scores \cite{gallegos2024bias,bai2025explicitly,lou2026anchoring}. 
The result supports a limited claim that the model associates some terms, groups or outcomes more strongly than others under the chosen operationalization. 
This claim remains close to the measured quantity, but still requires justification that the measured association is a meaningful proxy for the intended construct. 
By itself, it does not 
show how the model will behave in interaction, whether the association will cause harm, or whether it reflects a human-like cognitive mechanism \cite{gawronski2022implicit,jacobs2021measurement}.

\vspace{-0.25cm}\paragraph{Observable behaviour} refers to bias expressed in model outputs under specific evaluation conditions, like stereotype-consistent generations, framing-sensitive answers, or decision changes after biased cues \cite{gallegos2024bias,echterhoff-etal-2024-cognitive}. Observations are tied to the prompting setup, model version, decoding choices, and context \cite{hida-etal-2025-social,Holtzman2020The}.
This supports a context-bound behavioural interpretation in which the model produces biased outputs in the tested setting. 
This interpretation requires attention to the evaluation context, because outputs are shaped by prompts, model versions, decoding choices, refusal behaviour and interaction context in which the behaviour was observed \cite{hida-etal-2025-social,Holtzman2020The}. 
Stronger behavioural claims therefore require evidence that the pattern is robust across relevant prompts, models, decoding settings or task variants.

\vspace{-0.25cm}\paragraph{Normative harm} refers to model behaviour connected to individual or social consequences. 
Here, the interpretation moves beyond the existence of a pattern to whether it might 
reinforce inequality, degrade decisions or adversely affect users, groups or institutions in a particular context. 
For example, demographic disparity in a simulated hiring recommendation may matter because similar behaviour in deployed LLMs could reinforce unequal treatment \cite{blodgett2020language}. Harm claims require more than detecting a pattern. They require an account of the setting, affected users/groups, and consequences that follow from the model output \cite{blodgett2020language,gallegos2024bias}.

\vspace{-0.2cm}\paragraph{Psychological analogy} refers to treating model behaviour as meaningfully similar to a human cognitive or social-cognitive bias \cite{bai2025explicitly,echterhoff-etal-2024-cognitive,lou2026anchoring}. 
This is a stronger interpretation than an output pattern as it treats the model behaviour as understandable through a human-derived construct such as 
implicit bias, stereotype activation, anchoring, or confirmation bias. 
Such claims may be useful when carefully formulated, but they require explicit theoretical justification. 
In particular, researchers should specify which aspects of the human construct are preserved, absent, or transformed, and why the model-side observable supports the analogy \cite{gawronski2022implicit}.

The framework can be used as a reporting and interpretation checklist. It asks researchers to make explicit: 
(i) the human-derived construct being invoked; 
(ii) the model-side observable used to operationalize it;
(iii) the evaluation context in which the observable is produced; and
(iv) the strongest type of claim the design can support. 
As a result, the framework may leave the empirical measurement unchanged while changing the scope and strength of the conclusion drawn from it.

The framework does not make the classification of bias claims fully objective. Claims about harm and psychological analogy necessarily involve theoretical and contextual judgment. 
Its purpose is instead to make the basis for such judgments explicit and comparable across studies. 
Researchers should begin from the model-side observable and identify the strongest interpretation for which the required support is available. 
If the evidence consists of likelihood contrasts, embedding similarities or association scores, the warranted interpretation is closest to distributional association. 
If the evidence consists of generated outputs, rankings, recommendations, classifications or answer shifts under specified evaluation conditions, it may support observable behaviour, especially when robustness across prompts, models, decoding settings or task variants is shown. 
If the claim concerns harm, the evaluation must additionally specify a use context, affected groups and plausible consequences. 
If the claim invokes a human-derived construct, the evaluation must justify the mapping between the human construct and the model-side observable.

The IAT-inspired case provides a concrete example of how the framework changes the interpretation of an existing evaluation. 
\citet{bai2025explicitly} show that LLMs can appear explicitly unbiased in direct responses while still exhibiting biased associations under association-based measures. 
Our framework does \textit{not dispute this empirical finding, but it refines the conclusion that follows from it}. If the measured evidence is a likelihood or association difference between stereotype-consistent and -inconsistent group--attribute pairings, the most direct interpretation is \textit{distributional association}. 
Interpreting the same results as biased model \textit{behaviour} requires additional evidence that the associations shape generated outputs, ranking recommendations or other task behaviours.
Interpreting it as a \textit{normative harm} requires connecting the pattern to a specific use context, affected groups and plausible consequences. 
Interpreting it as human-like \textit{implicit bias} requires a \textit{psychological analogy} between the model-side observable and the human construct. 
The framework therefore changes the conclusion from a broad claim that the model has implicit bias to a more precise claim about what the evidence supports and what additional support would be needed for stronger interpretations. 

These interpretations 
mark different inferential endpoints, including what the model seems to associate, what it produces under evaluation conditions, what consequences that behaviour may have in context, and whether it is meaningfully analogous to a human cognitive or social-cognitive bias\footnote{Appendix~\ref{app:example} illustrates the same logic with an anchoring-style evaluation}.

\vspace{-0.3cm}\section{Conclusions}\vspace{-0.3cm}
Human-derived bias constructs offer valuable resources for evaluating LLMs beyond aggregate performance metrics and overt bias queries. 
However, these constructs do not transfer to LLM evaluation unchanged. When operationalized through token probabilities, generated outputs, rankings or answer shifts, the relation between evidence and interpretation becomes indirect. 
This paper argued that such transfer creates three sources of mismatch: construct, subject and evaluation-context. Mismatches do not invalidate psychological paradigms as tools for LLM evaluation, but show why claims about model bias should be stated more precisely. By distinguishing distributional association, observable behaviour, normative harm and psychological analogy, our framework shapes bias evaluation not only as a measurement task, but also as a problem of scientific interpretation. As LLMs become embedded in 
daily life tasks, NLP research needs not only better probes, but also clearer accounts of what those probes can and cannot reveal.

\section*{Limitations}
We acknowledge a number of limitations. 
First, it is a conceptual and literature-based contribution rather than an empirical evaluation. We do not test whether specific LLM bias probes produce stable results across models, prompts, languages or deployment settings. Instead, our goal is to clarify the kinds of interpretations that such probes can support. Future work could empirically examine how existing evaluations move between distributional, behavioural, normative and psychological claims, and whether those claims are warranted by the evidence provided.
Second, the paper focuses on two broad families of human-derived constructs, social-cognitive bias constructs and cognitive biases in judgment and reasoning. This scope does not include all uses of psychological concepts in LLM evaluation, such as personality, emotion, moral judgment, theory of mind or pragmatic reasoning. The analytical framework may be useful for these areas as well, but additional work would be needed to adapt it to constructs with different theoretical histories, operationalizations and evidential standards.
Finally, the four interpretations proposed here are not intended as an exhaustive taxonomy of model bias. Distributional association, observable behaviour, normative harm, and psychological analogy capture recurring ways in which bias-related findings are interpreted, but other distinctions may be needed for specific domains, modalities or applications. In particular, normative harm depends on social context, affected communities and deployment conditions, which cannot be fully specified by an abstract analytical framework alone.

\subsection*{Ethical Considerations}
This paper does not introduce new datasets, model evaluations, or human-subject experiments. Its ethical concern is therefore primarily interpretative. One risk of imprecise claims about LLM bias is that they may anthropomorphize models, obscure the mechanisms through which harms arise, or overstate what particular evaluations demonstrate. Conversely, overly narrow interpretations may understate risks when biased outputs are likely to affect users, groups or society in deployment settings.

A further ethical consideration concerns the use of human-derived constructs such as implicit bias, stereotype activation or cognitive bias. These terms carry theoretical and social significance from psychology and social science. When applied to LLMs, they should be used in ways that avoid misleading equivalences between human cognition and model behaviour. Clearer distinctions between distributional association, observable behaviour, normative harm and psychological analogy can support more responsible communication of bias findings to researchers, developers, policymakers and affected communities.

\subsection*{Acknowledgments}
This research was funded in whole or in part by the Austrian Science Fund (FWF): \href{https://doi.org/10.55776/COE12}{10.55776/COE12}.

\subsection*{Declaration of Use of Generative AI Tools}
Generative AI tools were used solely to assist with grammar correction and language polishing; all conceptual content, analysis, and final decisions remain the authors' responsibility.

\bibliography{custom}

\newpage

\appendix

\section{Examples of the Inferential Gap in Existing Bias Evaluations}\label{app:existing-examples}

This Appendix provides examples of how the inferential gap discussed in the paper appears in existing LLM bias-evaluation studies. The goal is \textit{not to dismiss these studies or dispute their empirical findings}. Rather, the examples show \textit{how human-derived constructs can support useful measurements while also inviting stronger interpretations that require additional support}.

\paragraph{Political bias and self-perception.}
\citet{https://doi.org/10.1155/2024/7115633} evaluate ChatGPT using political-compass questionnaires, political-affiliation questions and personality-related instruments. 
These instruments provide evidence about how the model responds under particular questionnaire conditions. However, stronger interpretations, such as claims that the model holds political views, perceives itself as having certain personality traits, or even that it has a stable personality type, depend on whether human-oriented political and personality instruments can validly measure such constructs in LLMs. 
This illustrates construct mismatch and subject mismatch, as political identity, self-perception and personality are human constructs involving social, experiential and psychological dimensions.

\paragraph{Social desirability bias.}
\citet{salecha2024large} study social desirability bias in LLM responses to Big Five personality surveys. 
Their findings show that model responses shift toward socially desirable trait profiles when the evaluation context becomes identifiable.
This is evidence about response behaviour under a psychometric evaluation setting. 
At the same time, interpreting the pattern as human-like social desirability bias requires specifying which parts of the human construct are preserved or transformed in the LLM setting.
In humans, social desirability is often linked to self-representation, impression management and sensitivity to social norms. 
In LLMs, similar response patterns may instead reflect training data, instruction following, alignment procedures or learned conventions. 
 
\paragraph{Anchoring bias.}
\citet{takenami-etal-2025-cognitive} examine anchoring effects in LLM-based price negotiation simulations. Results show that model outputs and negotiation outcomes shift under anchoring-style interventions. This provides evidence of context-sensitive behaviour in a simulated negotiation setting. 
A stronger interpretation that LLMs are influenced by anchoring in the same sense as humans requires an explicit mapping between the human cognitive-bias construct and the model-side behaviour.

These examples show why the inferential gap is not merely hypothetical. Existing evaluations can produce informative findings while still requiring care in how those findings are translated into claims about model beliefs, personality, social desirability, cognitive bias, harm or human-like psychological processes.

\section{Example Distinguishing Two Families of Bias Probes}\label{app:bias-probe-families}

This Appendix illustrates the distinction between the two families of bias probes discussed in Section~\ref{sec:evaluation-human-bias}. 
Both families transfer human-derived constructs to model-side observables, but they differ in the construct invoked, the evidence collected and the interpretation that is most directly warranted.

\paragraph{Social-cognitive bias probe.}
An IAT-inspired probe tests whether an LLM produces stronger associations between social groups and stereotyped attributes, for example whether gendered terms are more strongly associated with career- or family-related terms. 
The human-derived construct comes from social cognition and prejudice research, where the IAT uses response-time differences to infer relative associations between concepts \cite{greenwald1995implicit,greenwald1998measuring}.
In an LLM adaptation, the model-side observable may instead be a likelihood contrast, embedding similarity, ranking preference or generated continuation.
Therefore, the most direct interpretation is a claim about \textit{distributional association}, unless further evidence shows that the association affects generated outputs, produces harms in a specific use context or is analogous to human implicit bias.

\paragraph{Cognitive-bias probe.}
An anchoring-style probe asks whether an LLM's answer shifts toward an irrelevant numerical cue. The human-derived construct comes from work on judgment and reasoning, where anchoring refers to the influence of an initial value on subsequent estimates or decisions \cite{tversky1974judgment}. In an LLM adaptation, the model-side observable may be an answer shift across controlled prompt variants, such as different estimates after a low or high anchor. 
The most direct interpretation is a claim about \textit{observable behaviour} under the tested prompt conditions. 
A stronger claim about human-like anchoring requires showing that the anchor is task-irrelevant, that the task has a meaningful target or baseline, and that the model systematically shifts toward the anchor across controlled variants.

This contrast shows why the two probes should not be collapsed into a single notion of \textit{LLM bias}. 
In the first case, the central evidential issue is whether a model-side association is a meaningful proxy for a social-cognitive construct. Whereas in the second case, the central issue is whether an output shift reflects a cognitive-bias pattern rather than general prompt sensitivity. 
In both cases, stronger claims about harm or psychological analogy require additional support beyond the initial model-side measurement. 

\section{Applying the Framework to an Anchoring-Style Evaluation}\label{app:example}

Consider an evaluation in which a model gives a higher estimate after exposure to an arbitrary numerical anchor. The direct finding is that the model output shifts under a controlled prompt variation. Under our framework, this supports an \textit{observable behaviour} claim, namely that the model is sensitive to an anchoring-style cue under the tested conditions. 

A stronger claim about human-like anchoring requires additional support. Researchers should specify what is meant by anchoring in the LLM setting, why the anchor is task-irrelevant, whether the task has a meaningful target answer or baseline, and whether the model systematically shifts toward the anchor across controlled variants. Without this mapping, the result is better described as context-sensitive model behaviour than as human-like anchoring.

A harm-based interpretation would require further support of a different kind. The output shift would need to be connected to a concrete use setting, affected users or groups, and plausible consequences, such as distorted salary estimates in compensation support, biased evaluations in financial decision support or unfair outcomes in online marketplaces. 
Thus, the framework changes the conclusion from a broad claim that the model has anchoring bias to a more precise claim about context-sensitive model behaviour, unless additional evidence supports the psychological or harm-based interpretations.


\end{document}